\documentclass[pdflatex,sn-mathphys-num]{sn-jnl}

\usepackage{graphicx}%
\usepackage{multirow}%
\usepackage{amsmath,amssymb,amsfonts}%
\usepackage{amsthm}%
\usepackage{mathrsfs}%
\usepackage[title]{appendix}%
\usepackage{xcolor}%
\usepackage{textcomp}%
\usepackage{manyfoot}%
\usepackage{booktabs}%
\usepackage{algorithm}%
\usepackage{algorithmicx}%
\usepackage{algpseudocode}%
\usepackage{listings}%
\usepackage{hyperref}%
\usepackage{minted}%
\usepackage{array}%
\usepackage{pdfpages}%
\usepackage{pifont}%
\usepackage{textcomp}%
\usepackage{pdflscape}%
\usepackage{array}%
\usepackage{float}%
\usepackage{caption}%
\usepackage{subcaption}%
\usepackage{tabularx}%
\usepackage[utf8]{inputenc}%
\usepackage{siunitx}
\usepackage{placeins}
\theoremstyle{thmstyleone}%

\theoremstyle{thmstyletwo}%

\theoremstyle{thmstylethree}%

\unnumbered

\begin{document}

\title[Article Title]{The Bundesrat corpus: A dataset of political discourse in the German Federal Council}

\author*[1]{\fnm{Adama} \sur{Njie}}\email{njie@sustainability.rwth-aachen.de}

\author[1]{\fnm{Leandro} \sur{Laperne}}

\author[1]{\fnm{Laksh} \sur{Agarwal}}

\author[1]{\fnm{Ali} \sur{Torkayesh}}

\author[1]{\fnm{Sandra} \sur{Venghaus}}

\affil[1]{\orgdiv{Economics of Sustainability}, \orgname{School of Business and Economics, RWTH Aachen University}, \orgaddress{\postcode{52072}, \city{Aachen}, \country{Germany}}}

\abstract{The Bundesrat is Germany’s second legislative chamber and a constitutional veto player within its political system. The members of the Bundesrat are delegates of state governments. Tracing the political discourse in the Bundesrat is challenging as it has a limited machine-readable corpus available. To address this issue, we present a speaker-attributed corpus of \num{1067} Bundesrat plenary sessions from the chamber’s first session on 7 September 1949 to 10 July 2026. In this period, a total of \num{53362} speeches by \num{2473} individuals were delivered in the chamber, including written statements. Each speech includes a speech ID, the full German text, the speaker's name, the date and location of the session, the federal state that the speaker represents, the speaker's role (if applicable), the type of contribution, and the speaker's gender. The dataset provides interdisciplinary research opportunities across computational political science, legislative studies, and  parliamentary discourse analysis, coupled with discourse from the Bundestag.}


\maketitle

\section{Background \& Summary}\label{sec1}
Parliamentary discourse has become a primary source of evidence in political science, and the methods that read it including scaling, classification, emotion detection, group-difference and measurement depend on corpora that are complete, attributed and machine-readable \cite{slapin2014,proksch2015politics,back2021,gentzkow2019,rheault2016,peterson2018}. That dependency has driven a wave of national corpus releases, several of which are published in Scientific Data, including Norway \cite{lapponi2018talk,fiva2025norwegian}, Australia \cite{katz2023digitization}, Finland \cite{simola2025finnish}, Italy \cite{cova2025new}, and cross-national collections covering the key legislative chambers of nine democracies \cite{rauh2020parlspeech}. For Germany, there exist several datasets on the German Bundestag built on the open access data from \cite{bundestag_plenarprotokolle}, including A Parliamentary Discourse Dataset from the German Bundestag
\cite{njie_2026_21258818}, GermaParl \cite{blatte2018germaparl,blaette2023germaparl} and Open Discourse \cite{richter2023open}, and its German state parliaments as SpeakGer \cite{lange2023speakger}. 

The second legislative body in the German government is the Bundesrat that acts as the second legislative chamber of the country. The Bundesrat holds an absolute veto over the category of federal legislation requiring its consent, which has historically covered a substantial share of some bills \cite{Reuter2020}. The chamber is structurally different than the Bundestag as its members are not directly elected rather they are appointed delegates of the federal state governments \cite{Reuter2020}. To understand the legislative difference between the Bundestag and Bundesrat consider following example, where the Bundestag corpus answers "what does the SPD utter?" whilst the Bundesrat corpus corresponds to "what does North Rhine-Westphalia say?", making territorial representation, Land-versus-party loyalty, and federal state conflict directly measurable.

Despite significant importance of the political discourse within the Bundesrat in the national politics, to the best of authors' knowledge, there exists no corpus dataset of the Bundesrat. The \num{1067} plenary sittings since 1949 exist only as PDF documents on the chamber's official website. The 1949–-1999 portion of the corpus includes roughly \num{27000} pages that exists only as scanned images, with no text layer at all. To address the aforementioned data gap, in this Data Descriptor, we present a complete speaker-attributed corpus of \num{1067} Bundesrat plenary sessions from the chamber's first sitting on 7 September 1949 to 10 July 2026. It comprises \num{53362} speeches delivered in the chamber including written statements. What makes this corpus standout is the fact it is indexed territorial-based rather than partisan, this provides researchers and other interested groups a different perspective to German political discourse. 


The remainder of this paper is structured as follows: The Methods section describes the extraction, cleaning, and speaker attribution pipeline. The Data Records section documents the schema and corpus overview. The Technical Validation section reports quality checks. The Usage Notes section provides recommended use of the dataset.

\section{Methods}\label{sec2}

\subsection{Institutional Settings}
Unlike Bundestag where parties structure discourse, the Bundesrat is organized around territorial representation. Members of the Bundesrat are delegates of the federal governments rather than representatives elected to the chamber. Therefore, the chamber has no fractions, a delegate speaks for a state and not for a parliamentary party. Article 51(3) of The Basic Law for the Federal Republic of Germany (GG) \cite{grundgesetz} requires each federal state to cast its votes uniformly. Additionally, Article 51(2) GG \cite{grundgesetz} sets each state's number of votes by population, ranging from three to six.  As a permanent body of the government, it is never dissolved and holds no elections, and its sittings have been numbered continuously since 1949 rather than being grouped into legislative periods. Moreover, Article 52(1) GG \cite{grundgesetz} states that the Bundesrat elects its president each year based on the Königsteiner Agreement of 1950, which ensures that the office passes among the state premiers (Ministerpräsidenten) in descending order of state population. 

\subsection{Data Source}
All source documents are the official \textit{Plenarprotokolle} published by the Bundesrat and retrieved from its plenary protocol archive at \url{https://www.bundesrat.de/DE/service/archiv/pl-protokoll-archiv/pl-protokoll-archiv-node.html}. It is published in different period-based web pages (e.g., 1949--1950, 1951--1955, ..., 2026--2030) as PDF documents. Each PDF document represents one session and the filename carries useful information, including session number and date that are essential for building the metadata. The corpus is designed to be extensible; thus, newer sessions can be added accordingly. This feature is used to add the 2026 sessions, including session numbers \num{1061}, \num{1062}, \num{1063}, \num{1064}, \num{1065}, \num{1066}, and \num{1067}.

\subsection{Data Preprocessing}
The data preprocessing pipeline is designed in five stages, including text extraction, cleaning, speaker attribution, dataset creation, and persons resolution. Each stage is resume-safe and idempotent, that is to say, it checks whether the session's output file already exists and it skips it if it does. Re-running after a crash or a change in some files picks up where it stopped or where the changes happened. 

The text extraction stage uses four extraction methods, which are applied in order of decreasing fidelity. The \textit{pdftotext -layout} method preserves the two columns structure from embedded text layer  as Bundesrat documents have two columns. The \textit{pdftotext -raw} method uses the same layer, but in content stream order for older files. The \textit{pdfminer} method serves as a pure-Python fallback for files when \texttt{pdftotext} (poppler) fails to parse. The last resort is \textit{Tesseract OCR (German)} if all the previous methods fail. If pages show no text layer at all, optical character recognition (OCR) renders the page as an image and infers characters from pixels which may occasionally introduce wrong or weird characters. There are two physical categories of PDFs that are handled in this pipeline. The first category is the image scans of 747 sessions from 1949 to 1999. The PyMuPDF turns each page into a bitmap image and the \texttt{Tesseract} then recognizes the characters in the image using the \texttt{German model (deu)}. The other category is the born-digital PDF document with 320 sessions from 2000 to 2026 parsed with the \texttt{pdftotext}.


The text cleaning stage consists of 14 steps, ordered according to dependencies rather than convenience. The extracted raw text is not just speech. It also includes artifacts of parliamentary protocol, such as running headers, footnote markers, page numbers, tables of contents, and attendance rosters. This stage removed apparatuses whilst leaving the speech untouched. Table \ref{tab:14-cleaning-steps} summarizes the cleaning steps and the rationale behind each of them.

\begin{table}[htbp]
\centering
\caption{The text-cleaning has 14 operations applied in a fixed order; \texttt{remove\_headers} and \texttt{normalize\_whitespace} run a second time after hyphen rejoining, which exposes residual headers previously split across a line break.}
\label{tab:14-cleaning-steps}
\small
\begin{tabularx}{\textwidth}{@{}r l X@{}}
\toprule
\# & Operation & Removes or repairs \\
\midrule
1  & \texttt{fix\_column\_order}              & Two-column block order (ACBD $\rightarrow$ ABCD). \\
2  & \texttt{remove\_page\_breaks}            & Page-break markers inserted at extraction. \\
3  & \texttt{remove\_cid\_artefacts}          & \texttt{(cid:\textit{N})} font artefacts. \\
4  & \texttt{remove\_headers}                 & Running page headers (\textit{Bundesrat -- N. Sitzung -- date}). \\
5  & \texttt{remove\_running\_speaker\_headers} & Speaker name reprinted at the top of a column. \\
6  & \texttt{remove\_toc}                     & Table of contents and attendance roster. \\
7  & \texttt{remove\_page\_numbers}           & Standalone page numbers. \\
8  & \texttt{remove\_footnotes}               & Footnote markers and colophon lines. \\
9  & \texttt{remove\_margin\_markers}         & Column markers (A)--(D), repaired by position. \\
10 & \texttt{remove\_bare\_separators}        & Rule and dash artefacts on otherwise empty lines. \\
11 & \texttt{remove\_backm\_references}       & Back matter.  \\
12 & \texttt{normalize\_whitespace}           & Whitespace runs, spacing before punctuation. \\
13 & \texttt{remove\_duplicated\_blocks}      & Adjacent repeated blocks from column reconstruction. \\
14 & \texttt{rejoin\_hyphens}                 & Words broken across a line or column break. \\
\bottomrule
\end{tabularx}
\end{table}

The speaker attribution in the Bundesrat has structural differences from the \mbox{Bundestag}. Aside from the procedural chairs and other roles, a typical Bundesrat speech has a \texttt{federal state} in parenthesis. A block of text is considered a speech if the text is attributed to a person based on patterns. The pattern has to be terminated by a colon followed by a block of text. Interjections from other members or the chair are not included in the speaker's text. The patterns are broadly divided into five families; chair, state delegate, rapporteur, Federal official, and Bundestag figure. It is important to note that surnames are set in capitals in 96--98\% of speeches before 1960 and in 0\% afterwards, a change independent of the 2000 boundary between scanned and born-digital sources. Details regarding the speakers are summarized in Table \ref{tab:speaker-attribution}.

\newcommand{\att}[1]{{\ttfamily\hyphenchar\font=`\-\relax #1}}

\begin{table*}[htbp]
\centering
\caption{Speaker-attribution families and their two typographic variants. Examples are drawn unmodified from the corpus.}
\label{tab:speaker-attribution}
\footnotesize
\begin{tabularx}{\textwidth}{@{}l l X l r@{}}
\toprule
Family & Era & Attribution form & Fields & Speeches \\
\midrule
Chair               & 1949--59   & \att{Präsident Dr. EHARD:}                                 & name, role       & \num{23141} \\
                    & 1960--2026 & \att{Präsident Klaus Wowereit:}                            &                  &             \\
\addlinespace
State delegate      & 1949--59   & \att{RENNER (Württemberg-Hohenzollern):}                   & name, land       & \num{15678} \\
                    & 1960--2026 & \att{Hans Kaiser (Thüringen):}                             &                  &             \\
\addlinespace
\quad as rapporteur & 1949--59   & \att{KUBEL (Niedersachsen), Berichterstatter:}             & name, land, role & \num{4061}  \\
                    & 1960--2026 & \att{Dr. Heinsen (Hamburg), Berichterstatter:}             &                  &             \\
\addlinespace
Federal official    & 1949--59   & \att{SCHÄFFER, Bundesminister der Finanzen:}               & name, role       & \num{3924}  \\
                    & 1960--2026 & \att{Matthöfer, Bundesminister der Finanzen:}              &                  &             \\
\addlinespace
Bundestag figure    & 1949--59   & \att{Bundestagsabgeordneter Dr. MENZEL, Berichterstatter:} & name, role       & \num{114}   \\
                    & 1960--2026 & \att{Bundestagsabgeordneter Westphal, Berichterstatter:}   &                  &             \\
\bottomrule
\end{tabularx}
\end{table*}

\subsection{Derived Attributes}
The Bundesrat does not have a dedicated biography register for members, similar to MDB\_Stammdaten; therefore, \texttt{gender} attribute is inferred from the corpus itself whilst \texttt{presiding\_land} attribute is derived from the corpus instead of an external source.

\begin{figure*}[htbp]
\centering
\includegraphics[width=\textwidth]{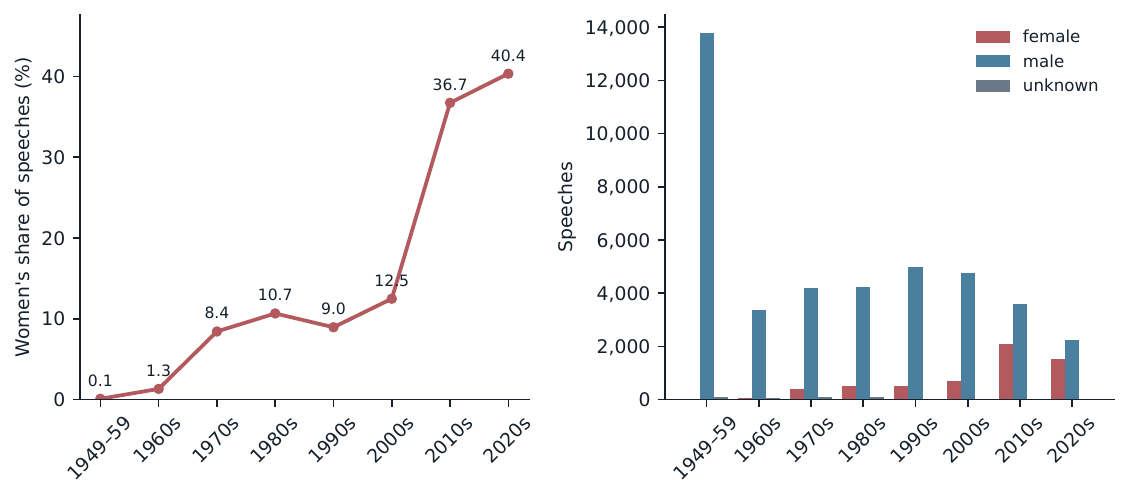}
\caption{Women in the Bundesrat between 1949--2026.}
\label{fig:gender}
\end{figure*}

To conduct downstream processes such as gender analysis, it is essential to know the gender of the speakers. In the German language, office nouns are inflected for gender and this is usually stated explicitly in the protocol. For example, Präsidentin, Bundesministerin, Berichterstatterin alongside the courtesy title Frau denotes a female speaker. In this context, the gender field is determined from the evidence on three tiers of decreasing directness and the chosen tier for each speaker is specified in the \texttt{gender\_source} column in the dataset. The first tier is the speaker's \texttt{own attribution line}, and it determined the genders of \num{1744} speakers. The second tier is the \texttt{given name} via \texttt{gender\_guesser} and accounts for 233 persons. The last tier is how other speakers refer to them in the body of the text and it accounts for 104 persons. In case all tiers fail, a person's gender is unknown and is left \texttt{blank} or \texttt{null}. 
The protocols predominantly identified male speakers at \num{1681} of the \num{2473} identified persons with a contribution rate of 84.9\% of the speeches. The number of female speakers identified by the protocol's own form of address is 400, and accounts for 13.2\% of the speeches whilst 392 persons gender could not be determined. Figure \ref{fig:gender} shows the progression of the proportion of female speech across the history of the Bundesrat.


The \texttt{presiding\_land} attribute is derived to determine the federal states of the presiding officers as in the Königsteiner Agreement. Since the presiding officer represents no state while speaking, and accounts for 49\% of all speeches in the corpus, the dataset records \texttt{presiding\_land}, the state holding the presidency at each sitting, which lifts territorial attribution from 41.7\% to 99.3\%. This derivation produces historical sequence of the Bundesrat presidency as shown in Table \ref{tab:presidency}.

\begin{table}[htbp]
\centering
\caption{The first and last five of the 77 presidency terms.}
\label{tab:presidency}
\small
\begin{tabular}{@{}llllrr@{}}
\toprule
\# & Federal state & Presiding chair & Term & Sessions & Days \\
\midrule
1 & Nordrhein-Westfalen & Arnold     & Sep 1949 -- Aug 1950 & 32 & 338 \\
2 & Bayern              & Dr.\ Ehard & Sep 1950 -- Sep 1951 & 34 & 364 \\
3 & Niedersachsen       & Kopf       & Sep 1951 -- Jul 1952 & 23 & 313 \\
4 & Württemberg-Baden   & Dr.\ Maier & Sep 1952 -- Jul 1953 & 23 & 308 \\
5 & Hessen              & Dr.\ Zinn  & Oct 1953 -- Jul 1954 & 13 & 266 \\
\addlinespace
\multicolumn{6}{c}{\textit{67 further terms, Oct 1954 -- Oct 2021, covering \num{867} sessions}} \\
\addlinespace
73 & Thüringen              & Bodo Ramelow            & Nov 2021 -- Oct 2022 & 17 & 357 \\
74 & Hamburg                & Dr.\ Peter Tschentscher & Nov 2022 -- Oct 2023 & 11 & 340 \\
75 & Mecklenburg-Vorpommern & Manuela Schwesig        & Nov 2023 -- Oct 2024 & 11 & 329 \\
76 & Saarland               & Anke Rehlinger          & Nov 2024 -- Oct 2025 & 10 & 329 \\
77 & Bremen & Dr.\ Andreas Bovenschulte & Nov 2025 -- Jul 2026 & 9 & 231 \\
\bottomrule
\multicolumn{6}{@{}l@{}}{%
  \footnotesize
  \parbox{\dimexpr\linewidth-2\tabcolsep}{%
    \textsuperscript{\dag} Term 77 was still in progress at the corpus cut-off of 10 July 2026, so its session and day counts are partial and not comparable with the completed terms.
  }%
} \\
\end{tabular}
\end{table}

\section{Data Records}\label{sec3}
\subsection{Dataset Overview}
The dataset contains four tables built from \num{1067} per session attribution files and the extraction catalog by a packaging step that performs no parsing of its own. Every field in the corpus is either carried from the attribution files or computed by aggregation so the package can be built at any time without disturbing the speech layer. 

The four main tables are in a dual format with identical content, which is deliberate. While the Parquet file is suitable for pipeline analysis and preserves data types, Parquet dependency is a real barrier for the use of dataset in different software packages. Hence, identical CSV files are included in the corpus. The relationship among tables in the corpus is described in the \texttt{datapackage.json} file as a frictionless data schema. Parquet files are Snappy compressed and written with PyArrow14. All text fields are UTF-8 encoded, null values are represented as \texttt{null} in Parquet and as empty strings in the CSV file. The dataset is publicly available on Zenodo \url{https://zenodo.org/records/22086439} under a Creative Commons Attribution 4.0 International licence (CC BY4.0). Table \ref{tab:data-files-overview} presents the files and what they constitute.

\begin{table}[htbp]
\centering
\caption{An overview of the Bundesrat dataset.}
\label{tab:data-files-overview}
\small
\begin{tabular}{@{}llrp{0.42\textwidth}@{}}
\toprule
File & Format & Rows & Description \\
\midrule
\texttt{speeches.parquet} / \texttt{.csv} & Parquet, CSV & \num{53362} &
  One row per attributed contribution by a named member; full text, speaker name
  and resolved person, role, represented state, and session identifiers.
  \texttt{contribution\_type} distinguishes the \num{47290} delivered in the
  chamber from the \num{6072} submitted to the record in writing. \\
\texttt{sessions.parquet} / \texttt{.csv} & Parquet, CSV & \num{1067} &
  One row per plenary session; date, meeting city, presiding state,
  extraction engine, and contribution counts. \\
\texttt{persons.parquet} / \texttt{.csv} & Parquet, CSV & \num{2473} &
  One row per resolved individual; gender, modal role and state, and
  aggregates across all of that person's name strings. \\
\texttt{speakers.parquet} / \texttt{.csv} & Parquet, CSV & \num{2907} &
  One row per unique speaker name string; gender, modal role and state,
  and activity span. \\
\texttt{laender.csv} & CSV & \num{19} &
  Reference table of the Bundesländer represented in the corpus; observed first
  and last year, counts, and whether the state is historical. \\
\texttt{datapackage.json} & JSON & --- &
  Frictionless data descriptor; column types, descriptions, and foreign-key
  constraints, declaring both formats. \\
\texttt{CITATION.cff} & YAML & --- &
  Machine-readable citation metadata (CFF v1.2; ingested by GitHub and Zenodo). \\
\texttt{README.md} & Markdown & --- &
  Dataset overview and Python quickstart code for loading each file. \\
\texttt{LICENSE} & Text & --- &
  Creative Commons Attribution 4.0 International (CC BY 4.0). \\
\bottomrule
\end{tabular}
\end{table}

The \texttt{speeches} table is the main table in the corpus; it contains all the identified speeches from the Bundesrat between 7 September 1949 to 10 July 2026. Each row represents an attributed speech, with speech\_id acting as the primary key, which is a composite of the session\_id and the sequence number of the speech within the dataset (\texttt{Session\_0001\_0001}). The table contains \num{53362} individual speeches, full column definitions, and other relevant information as shown in Table \ref{tab:cols-speeches}.

\begin{table}[htbp]
\centering
\caption{Column definitions for \texttt{speeches} table.}
\label{tab:cols-speeches}
\small
\begin{tabular}{@{}llp{0.56\textwidth}@{}}
\toprule
Field & Type & Description \\
\midrule
\texttt{speech\_id}    & string & Primary key, e.g. \texttt{Session\_0001\_0001}. \\
\texttt{session\_id}   & string & External key $\rightarrow$ \texttt{sessions.parquet}. \\
\texttt{sequence}      & int    & Position within session. \\
\texttt{date\_iso}     & date   & Session date (denormalized). \\
\texttt{year}          & int    & Calendar year (denormalized). \\
\texttt{meeting\_city} & string & Sitting location (denormalized). \\
\texttt{name}          & string & Normalized speaker name; surname-only in most pre-2000 sessions, full names from 2000. \\
\texttt{person\_id}    & string & External key $\rightarrow$ \texttt{persons.parquet}; null where identity could not be resolved with confidence. \\
\texttt{person\_name}  & string & Canonical display name of the resolved person. \\
\texttt{speaker\_raw}  & string & Raw attribution string from the source PDF. \\
\texttt{role}          & string & Speaker role; null for plain state delegates \\
\texttt{land}          & string & State (Bundesland) represented; null for the presiding chair and federal officials. \\
\texttt{contribution\_type} & string & How the contribution reached the record: \textit{spoken} (\num{47290} rows) delivered in the chamber, or \textit{written} (\num{6072}) submitted to the record. \\
\texttt{text}          & string & Full speech text in German. \\
\texttt{char\_offset}  & int    & Character offset of speech start in the cleaned source text file. \\
\bottomrule
\end{tabular}
\end{table}

Next, the \texttt{sessions} table describes each plenary sitting and each row represents a sitting. It generates most of the metadata information. The primary key in this table is the \texttt{session\_id} which is just the session number. It includes useful metadata such as the date on which the session was held and the city in which it took place. There are a total of \num{1067} sessions as of August 2026. It is important to note that session numbers 21, 28, 31, and 45 are marked as closed sessions with no published transcribe text. Table \ref{tab:cols-sessions} shows an overview of the \texttt{sessions} table.

\begin{table}[htbp]
\centering
\caption{Column definitions for \texttt{sessions} table.}
\label{tab:cols-sessions}
\small
\begin{tabular}{@{}llp{0.56\textwidth}@{}}
\toprule
Field & Type & Description \\
\midrule
\texttt{session\_id}         & string  & Primary key. \\
\texttt{session}             & int     & Continuous Bundesrat session number. \\
\texttt{date\_iso}           & date    & Session date. \\
\texttt{meeting\_city}       & string  & Bonn up to 1999, and Berlin from 2000. \\
\texttt{year}                & int     & Calendar year. \\
\texttt{source\_pdf}         & string  & Source PDF path, relative to the raw-PDF root. \\
\texttt{is\_schreibmaschine} & boolean & True, if a text comes from the typewritten rather than the printed protocol. \\
\texttt{md5}                 & string  & MD5 checksum of the source PDF. \\
\texttt{engine}              & string  & Extraction route: \texttt{pdftotext}, \texttt{pdftotext-raw}, \texttt{pdfminer}, or \texttt{ocr-tesseract}; flags OCR-derived sessions. \\
\texttt{presiding\_land}     & string  & State holding the presidency at this sitting. \\
\texttt{n\_speeches}         & int     & Speeches spoken in this session, recomputed at build time; counted on spoken contributions alone so that speeches per sitting hold same across the whole period. \\
\texttt{n\_written\_statements} & int  & Statements submitted to the record in writing. \\
\texttt{n\_documents\_excluded} & int  & Appendix items per sitting. \\
\bottomrule
\end{tabular}
\end{table}

The corpus carries speaker identity at two levels. The \texttt{speakers} provides an observational view of this dataset, where a row is recorded for every distinct string that appears in the protocol, along with the relevant aggregates and fulfilling the pattern requirements. It does not judge how a name is written or who anyone is. If a name is spelled three ways, there will be three rows. There are \num{2907} speaker records as shown in Table \ref{tab:cols-speakers}. The \texttt{persons} table is the resolved view. There is one record per individual after name strings have been group by procedures. There are \num{2473} \texttt{unique} speakers as shown in Table \ref{tab:cols-persons}.

\begin{table}[htbp]
\centering
\caption{Column definitions for \texttt{speakers} table.}
\label{tab:cols-speakers}
\small
\begin{tabular}{@{}llp{0.56\textwidth}@{}}
\toprule
Field & Type & Description \\
\midrule
\texttt{name}              & string & Primary key, name-string level: one person may appear under several variants, especially surname-only versus full name across the 2000 boundary. \\
\texttt{n\_speeches}       & int    & Contributions attributed in spoken and written formats. \\
\texttt{n\_sessions}       & int    & Distinct sessions. \\
\texttt{most\_common\_role}& string & Modal role; null if never assigned one. \\
\texttt{most\_common\_land}& string & Modal Bundesland (state); null for federal officials and presiding officers. \\
\texttt{first\_year}       & int    & Earliest year a speaker's name appears. \\
\texttt{last\_year}        & int    & Latest year a speaker's name appears. \\
\texttt{gender}            & string & \texttt{female} / \texttt{male} / null. \\
\texttt{gender\_source}    & string & \texttt{address}, \texttt{firstname}, \texttt{mention}, or null. \\
\texttt{person\_id}        & string & External key $\rightarrow$ \texttt{persons.parquet}; null if unresolved. \\
\bottomrule
\end{tabular}
\end{table}

\begin{table}[htbp]
\centering
\caption{Column definitions for \texttt{persons} table.}
\label{tab:cols-persons}
\small
\begin{tabular}{@{}llp{0.56\textwidth}@{}}
\toprule
Field & Type & Description \\
\midrule
\texttt{person\_id}         & string & Primary key: folded surname plus first year of appearance, e.g. \texttt{rau-1970}. \\
\texttt{person\_name}       & string & Canonical display name, the fullest name string observed. \\
\texttt{n\_speeches}        & int    & Contributions across all of this person's name strings. \\
\texttt{n\_sessions}        & int    & Distinct sessions in which a person appears. \\
\texttt{n\_name\_strings}   & int    & How many name strings were merged into a person. \\
\texttt{most\_common\_role} & string & Modal role across all contributions of a person. \\
\texttt{most\_common\_land} & string & Modal Bundesland (state); null for federal officials and presiding officers. \\
\texttt{first\_year}        & int    & Earliest year this person appears. \\
\texttt{last\_year}         & int    & Latest year this person appears. \\
\texttt{gender}             & string & \texttt{female} / \texttt{male} / null. \\
\texttt{gender\_source}     & string & \texttt{address}, \texttt{firstname} or \texttt{mention}; null where the protocols never mark it. \\
\bottomrule
\end{tabular}
\end{table}

An example of a distinction between \texttt{speakers} and \texttt{persons} tables is the case of Dr. Gebhard Müller, Minister-President of Württemberg-Hohenzollern and later Baden-Württemberg. Table \ref{tab:mueller-example} demonstrates this distinction clearly as six of the eight name strings are single-occurrence OCR corruptions of the given name. The seventh is the surname-only form used before full names became a standard format.

\begin{table}[htbp]
\centering
\caption{The two levels of speaker identity, for Dr.\ Gebhard Müller (Minister-President of Württemberg-Hohenzollern, later Baden-Württemberg).}
\label{tab:mueller-example}
\small
\begin{tabular}{@{}lrl@{}}
\toprule
\texttt{speakers} name string & Speeches & Years \\
\midrule
\texttt{Dr. Gebhard Müller}  & 258 & 1949--1955 \\
\texttt{Dr. Müller}          & 108 & 1949--1966 \\
\texttt{Dr. Gebbard Müller}  &   1 & 1950 \\
\texttt{Dr. Gbhard Müller}   &   1 & 1950 \\
\texttt{Dr. Gebbhard Müller} &   1 & 1949 \\
\texttt{Dr. Gebtard Müller}  &   1 & 1949 \\
\texttt{Dr. Gebtiard Müller} &   1 & 1953 \\
\texttt{Dr. Gerhard Müller}  &   1 & 1950 \\
\midrule
\texttt{persons} --- one individual & 372 & 1949--1966 \\
\bottomrule
\end{tabular}
\end{table}

\subsection{Territorial Composition}
\begin{figure}[htbp]
\centering
\includegraphics[width=\linewidth]{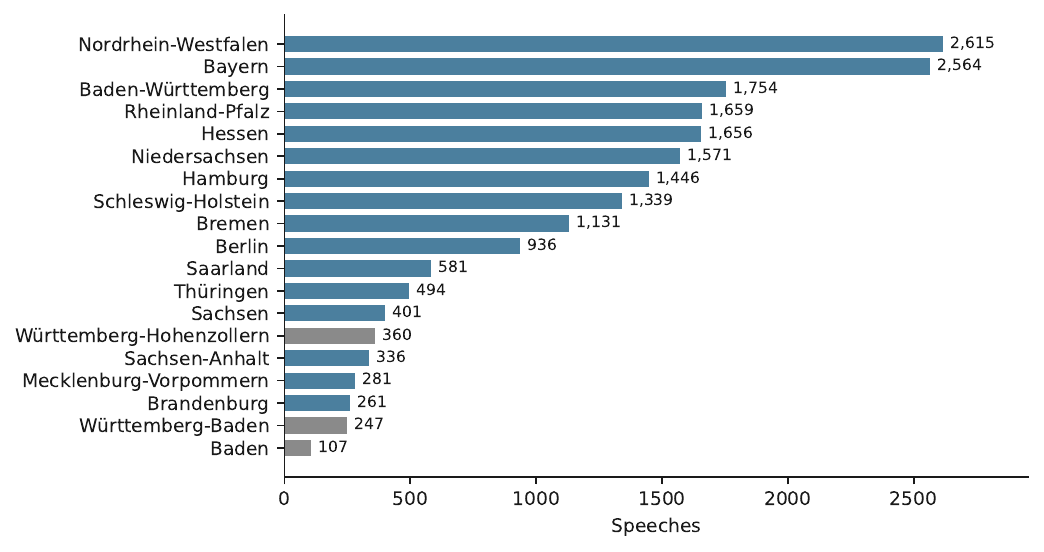}
\caption{Contributed speeches by representatives of federal states. The gray bars are the historical states that merged into Baden-Württemberg in 1952.}
\label{fig:land}
\end{figure}

Germany currently comprises 16 federal states, historically 19, before several were merged. The states represented in the Bundesrat are derived from attributions, thus, every transition of the states is recorded. Between 1949--1952, there were eleven states including Baden, Württemberg-Baden, and Württemberg-Hohenzollern. By 1952, these states merged into Baden-Württemberg. Saarland acceded as a state in 1957. Following the German reunification, Sachsen and Mecklenburg-Vorpommern, Sachsen-Anhalt, Brandenburg and Thüringen also joined the Bundesrat. To have an overview of the state-level contributions, Figure \ref{fig:land} illustrates the distribution of speeches across the individual federal states.

\section{Technical Validation}\label{sec4}
The corpus has been validated in four ways, three of which do not rely on the extraction pipeline that produced them. The Bundesrat protocols typically index their own speakers, and this independent index provides an external ground truth against which speaker recall can be measured directly. Each of these validation techniques addresses a different failure mode, and each can be reproduced using the codes released in the “Code Availability” section. 
The scripts \texttt{br\_05\_quality\_audit.py}, \texttt{br\_07\_toc\_recall.py}, and \texttt{br\_08\_error\_budget.py} produce all the tables and figures under this section.

\subsection{Corpus Completeness}
\begin{figure}[htbp]
\centering
\includegraphics[width=\linewidth]{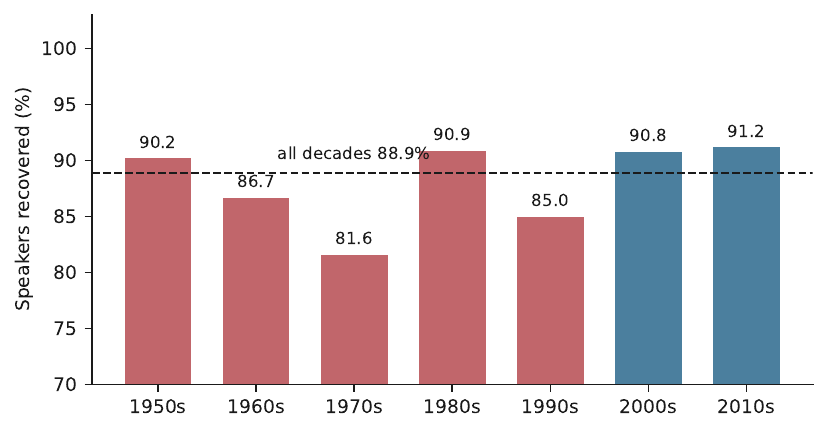}
\caption{Speaker-level recall against the printed table of contents by decade. The dashed line marks the corpus-wide figure. This is for spoken speakers only; written statements are excluded. The bars are colored by extraction route (scanned before 2000 and text layer after).}
\label{fig:recall}
\end{figure}

Each Plenarprotokoll document begins with an index showing all the speakers who contributed and the page on which they appear. As the index comes from the source document, it is an excellent tool for comparing extracted speakers with the material in the index. Of the \num{1067} Bundesrat sessions, 989 sessions yield an index that can be parsed. The remainder are the four closed sessions that the Bundesrat published no transcripts for. Additionally, those sessions whose index volumes are too damaged to be read. 
The speaker’s contribution is included in the appendix in specific cases where it was not delivered in the chamber, i.e., where a written document was submitted. This is usually denoted with an asterisk, for example, 384*D, where 384 is the page number and D denotes the bottom right side of the paper. Table \ref{tab:recall} and Figure \ref{fig:recall} illustrate corpus completeness and the recall rate of table of content.

\begin{table}[htbp]
\centering
\caption{Speaker-level recall against the protocols' printed index per decade.}
\label{tab:recall}
\small
\begin{tabular}{@{}lrrrrrr@{}}
\toprule
Decade & Indexed & Recovered & Missing & Anlagen & Unpaired & Recall \\
\midrule
1949--59 & \num{2327} & \num{2099} & \num{228} & \num{0}   & \num{59}   & 90.2\% \\
1960s    & \num{487}  & \num{422}  & \num{65}  & \num{0}   & \num{23}   & 86.7\% \\
1970s    & \num{848}  & \num{692}  & \num{156} & \num{4}   & \num{13}   & \textbf{81.6\%} \\
1980s    & \num{744}  & \num{676}  & \num{68}  & \num{63}  & \num{114}  & 90.9\% \\
1990s    & \num{747}  & \num{635}  & \num{112} & \num{59}  & \num{134}  & 85.0\% \\
2000s    & \num{1252} & \num{1137} & \num{115} & \num{534} & \num{137}  & 90.8\% \\
2010s    & \num{1299} & \num{1185} & \num{114} & \num{537} & \num{528}  & 91.2\% \\
2020s    & --         & --         & --        & \num{0}   & \num{1286} & -- \\
\midrule
\textbf{All} & \textbf{\num{7704}} & \textbf{\num{6846}} & \textbf{\num{858}} &
\textbf{\num{1197}} & \textbf{\num{2294}} & \textbf{88.9\%} \\
\bottomrule
\multicolumn{6}{@{}l@{}}{%
  \footnotesize
  \parbox{\dimexpr\linewidth-2\tabcolsep}{%
    \textit{Indexed} counts entries whose page reference carries no asterisk.
    \textit{Anlagen} are asterisked entries: statements submitted in writing, never delivered,
    and excluded from the denominator.
  }%
} \\
\end{tabular}
\end{table}

\subsection{Automated Quality Checks}
We conducted automated quality checks written in \texttt{br\_05\_quality\_audit.py}. The audit runs 15 checks on the dataset, and the code is included, so the validation can be reproduced rather than being a one-off claim. Eight of the 15 audit checks are at zero, including the following: page header inside a speech; column marker splitting a word; (cid:N) artefacts; null speaker name; stray field whitespace; and state outside the permitted 19 values. Table \ref{tab:audit} shows detail information on the quality of the checks performed and the state of each check. Moreover, the cleaning round substantially improved text fidelity in the speech corpus. Column markers splitting words were reduced from 1,674 cases to zero, while the total number of speeches containing column markers fell from 4,302 to 67. Page headers leaking into speeches were eliminated entirely, decreasing from 364 cases to zero. Overlength speeches caused by appendix absorption also declined markedly, from 198 to 8 cases.

\begin{table}[htbp]
\centering
\caption{The standing quality audit: 15 checks over all \num{53362} contributions, spoken and written, each matching a defect found during the dataset development.}
\label{tab:audit}
\small
\begin{tabularx}{\textwidth}{@{}Xrrl@{}}
\toprule
Check & Count & Baseline & Status \\
\midrule
\multicolumn{4}{@{}l}{\textit{Gated --- content}} \\
Page header inside a speech               & \num{0}  & \num{0}  & pass \\
Column marker splitting a word            & \num{0}  & \num{0}  & pass \\
\texttt{(cid:\textit{N})} font artefact   & \num{0}  & \num{0}  & pass \\
Unrejoined end-of-line hyphen             & \num{1}  & \num{1}  & pass \\
Attribution line inside a body            & \num{25} & \num{25} & pass \\
\addlinespace
\multicolumn{4}{@{}l}{\textit{Gated --- structure}} \\
Null or empty speaker name                & \num{0} & \num{0} & pass \\
Leading/trailing whitespace in a field    & \num{0} & \num{0} & pass \\
\texttt{land} outside the 19 Bundesländer & \num{0} & \num{0} & pass \\
\texttt{land} outside its own existence   & \num{0} & \num{0} & pass \\
Empty speech body                         & \num{0} & \num{0} & pass \\
\addlinespace
\multicolumn{4}{@{}l}{\textit{Informational --- not defects}} \\
Speaker's own name as a standalone line   & \num{1316} & --- & documented non-fix \\
Column marker anywhere                    & \num{84}   & --- & beside a capital or digit \\
Body under 20 characters                  & \num{138}  & --- & genuine procedural turns \\
Speech over \num{80000} characters        & \num{8}    & --- & residual appendix absorption \\
Body opens with the speaker's own name    & \num{0}    & --- & --- \\
\bottomrule
\end{tabularx}
\end{table}


\subsection{Metadata Completeness}
\begin{figure}[htbp]
\centering
\includegraphics[width=\linewidth]{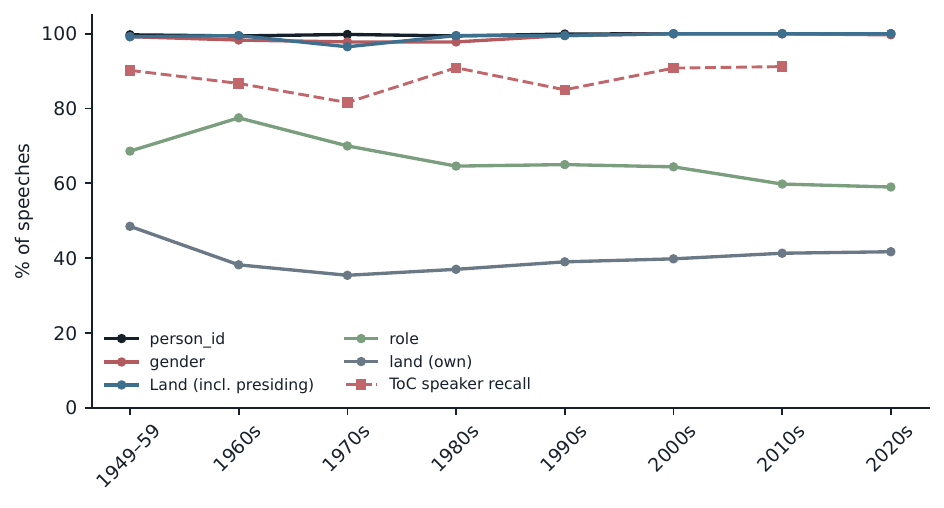}
\caption{Field completeness and measured recall by decade are shown on one axis. The dip in recall in the 1970s is not matched by any dip in the fields (\texttt{person\_id} 99.8\%, gender 97.8\%), indicating that the shortfall lies in scan quality rather than parsing. Land attribution moves with it (96.5\% against $\geq$99.2\% elsewhere), but draws on the same volumes, so corroborates this.}
\label{fig:completeness}
\end{figure}

Table \ref{tab:completeness} demonstrates the corpus completeness across seven decades over \num{53362} speeches. Also, Figure \ref{fig:completeness} demonstrates field completeness and measured recalls by each decade. 


\begin{table}[htbp]
\centering
\caption{Field completeness by decade over the \num{47290} spoken speeches.}
\label{tab:completeness}
\small
\begin{tabular}{@{}lrrrrrrr@{}}
\toprule
Decade & Speeches & \texttt{role} & \texttt{land} & +presiding & \texttt{person\_id} & gender & Recall \\
\midrule
1949--59 & \num{13895} & 68.6\% & 48.5\% & 99.2\%  & 99.7\%  & 99.2\%  & 90.2\% \\
1960s    & \num{3487}  & 77.5\% & 38.2\% & 99.5\%  & 99.4\%  & 98.3\%  & 86.7\% \\
1970s    & \num{4673}  & 70.0\% & 35.4\% & \textbf{96.5\%} & 99.8\% & 97.8\% & \textbf{81.6\%} \\
1980s    & \num{4848}  & 64.6\% & 37.0\% & 99.5\%  & 99.4\%  & 97.8\%  & 90.9\% \\
1990s    & \num{5496}  & 65.0\% & 39.0\% & 99.5\%  & 99.9\%  & 99.5\%  & 85.0\% \\
2000s    & \num{5437}  & 64.4\% & 39.8\% & 100.0\% & 100.0\% & 100.0\% & 90.8\% \\
2010s    & \num{5663}  & 59.8\% & 41.3\% & 100.0\% & 100.0\% & 100.0\% & 91.2\% \\
2020s    & \num{3791}  & 59.0\% & 41.7\% & 100.0\% & 100.0\% & 99.7\%  & -- \\
\midrule
\textbf{All} & \textbf{\num{47290}} & \textbf{66.3\%} & \textbf{41.7\%} & \textbf{99.3\%} & \textbf{99.8\%} & \textbf{99.1\%} & \textbf{88.9\%} \\
\bottomrule
\end{tabular}
\end{table}

\subsection{Topics Over Time as Validity Checks}
\begin{figure}[htbp]
    \centering

    \begin{subfigure}[t]{0.49\textwidth}
        \centering
        \includegraphics[width=\linewidth]{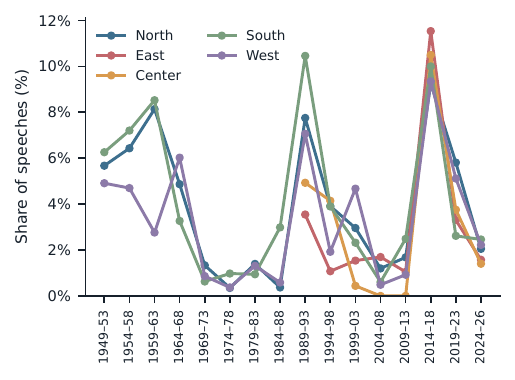}
        \caption{Refugees}
        \label{fig:kw_refugees}
    \end{subfigure}
    \hfill
    \begin{subfigure}[t]{0.49\textwidth}
        \centering
        \includegraphics[width=\linewidth]{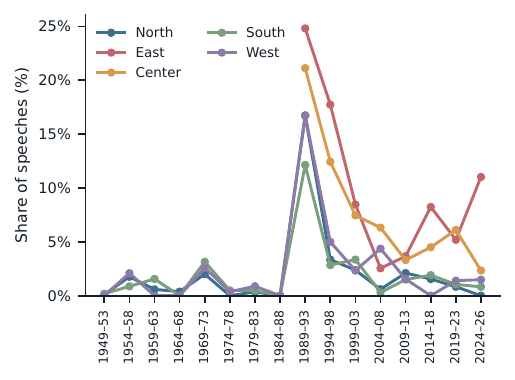}
        \caption{Reunification}
        \label{fig:kw_reunification}
    \end{subfigure}

    \vspace{0.5em}

    \begin{subfigure}[t]{0.49\textwidth}
        \centering
        \includegraphics[width=\linewidth]{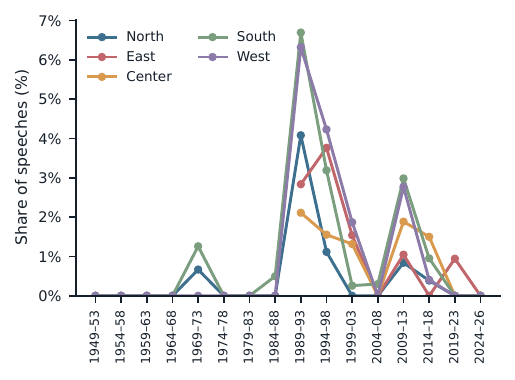}
        \caption{Monetary union, euro}
        \label{fig:kw_monetary}
    \end{subfigure}
    \hfill
    \begin{subfigure}[t]{0.49\textwidth}
        \centering
        \includegraphics[width=\linewidth]{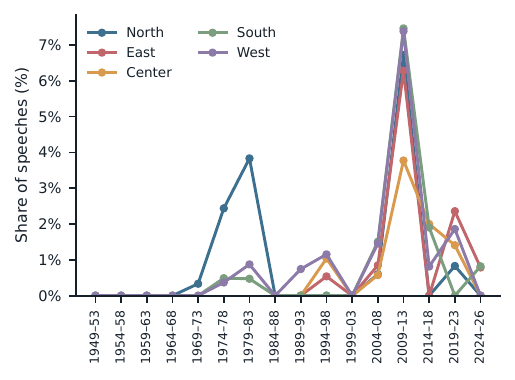}
        \caption{Financial crisis}
        \label{fig:kw_financial}
    \end{subfigure}

    \vspace{0.5em}

    \begin{subfigure}[t]{0.49\textwidth}
        \centering
        \includegraphics[width=\linewidth]{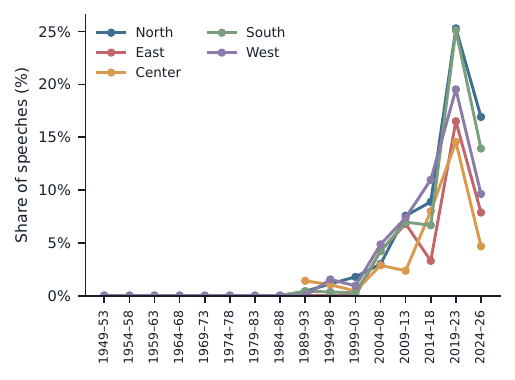}
        \caption{Climate policy}
        \label{fig:kw_climate}
    \end{subfigure}
    \hfill
    \begin{subfigure}[t]{0.49\textwidth}
        \centering
        \includegraphics[width=\linewidth]{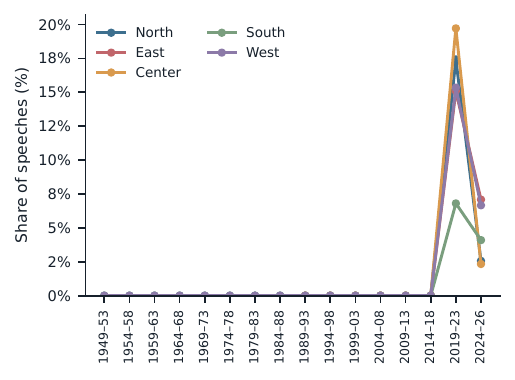}
        \caption{COVID-19}
        \label{fig:kw_covid}
    \end{subfigure}

    \caption{Event vocabulary by region, as a share of each region's spoken speeches in five-year periods.}
    \label{fig:keyword_trends_region}
\end{figure}

As a final validation step, in this part, we explored the speeches on key example events against what is curated in the corpus. We checked the dates of key historical events and matched it against the debates centred around such topics to determined if there is any alignment. Some of the key events that were checked to ensure they fall within the same timeline as the publicly available historical data include the following: flight and expulsion; German reunification; the introduction of the euro; financial crises; climate protection; and the outbreak of the Corona Virus (Covid-19). Further information regarding this validity check can be found in Table \ref{tab:events}. Figure \ref{fig:keyword_trends_region} shows the percentage of the states' speech share during these events. The states are grouped into \texttt{North} (Schleswig-Holstein, Hamburg, Bremen, Niedersachsen), \texttt{South} (Bayern, Baden-Württemberg and its predecessors Baden, Württemberg-Baden and Württemberg-Hohenzollern), \texttt{East} (Mecklenburg-Vorpommern, Brandenburg, Berlin, Sachsen), \texttt{West} (Nordrhein-Westfalen, Rheinland-Pfalz, Saarland), and \texttt{Center} (Hessen, Thüringen, Sachsen-Anhalt). 

\begin{table*}[tbp]
\centering
\caption{Event occurrence in the dataset by decade, as a share of spoken speeches in which the pattern occurs.}
\label{tab:events}
\small
\begin{tabular}{@{}lrrrrrrrrl@{}}
\toprule
Keyword group & 1949--59 & 1960s & 1970s & 1980s & 1990s & 2000s & 2010s & 2020s & Peak \\
\midrule
Refugees             & 4.55 & \textbf{5.53} & 0.51 & 0.91 & 2.55 & 0.92 & 3.44 & 1.87 & 1960s \\
Reunification        & 0.41 & 0.46 & 0.77 & 0.19 & \textbf{7.53} & 1.95 & 1.85 & 2.16 & 1990s \\
Monetary union, euro & ---  & ---  & 0.21 & 0.08 & \textbf{2.47} & 0.29 & 0.88 & 0.03 & 1990s \\
Financial crisis     & ---  & 0.03 & 0.24 & 0.33 & 0.20 & 0.92 & \textbf{1.43} & 0.50 & 2010s \\
Climate policy       & ---  & ---  & ---  & ---  & 0.31 & 1.53 & 5.14 & \textbf{8.89} & 2020s \\
COVID-19             & ---  & ---  & ---  & ---  & ---  & 0.02 & 0.02 & \textbf{7.83} & 2020s \\
\bottomrule
\end{tabular}
\end{table*}

\section{Usage Notes}\label{sec5}
The dataset offers numerous research opportunities in political science, computational politics, and the social sciences. Due to its 76-year span, the dataset is particularly well-suited to studying long-term ideological shifts and the role of different German states in political debates regarding a broad range of topics. Unlike other existing German parliamentary corpus, which are mainly focused on the Bundestag, the developed dataset provides answers to specific questions regarding political stance of different federal states rather than political parties. Furthermore, researchers can track speech behaviour and gender gaps in the Bundesrat and the rhetoric as the female representation steadily grow through the decades. 

The dataset contains two different forms of contribution. The first is the contribution of the members of the Bundesrat that was spoken in the chamber itself. These speeches are transcribed into PDF documents that our pipeline used. This accounts for \num{53362} unique speeches across seven decades and is represented in the \texttt{speeches} table. Every table is published as Parquet and as CSV with identical contents. Parquet preserves the data types and is substantially smaller in size compared to similar file types. However, CSV twin of each Parquet file also exists as Parquet data type is a challenging to be used for users of technical software packages, e.g., Stata, SPSS, and Excel.



\section{Data Availability}\label{sec6}
The dataset is available on Zenodo at \url{https://zenodo.org/records/22086439}.

\section{Code Availability}\label{sec7}
The code for the pipeline is available on GitHub at \url{https://github.com/adama-njie/bundesrat-dataset}.

\section{Author Contributions}\label{sec8}
\textbf{A.N.:} Conceptualization, formal analysis, methodology, resources, validation, data curation, writing - original draft, Writing - review \& editing. \textbf{L.L.:} formal analysis, methodology, resources, validation, data curation, writing - original draft, Writing - review. \textbf{L.A.:} formal analysis, data curation. \textbf{A.T.:} Conceptualization, formal analysis, resources, data curation, project administration, writing - original draft, Writing - review \& editing. \textbf{S.V.:} Conceptualization, resources, supervision, funding acquisition, Writing - review \& editing, and project administration.

\section{Acknowledgment}\label{sec9}
Laksh Agarwal’s affiliation with the Economics of Sustainability at the School of Business and Economics at RWTH Aachen University reflects his participation in a volunteer remote internship in 2025.

\section{Competing Interest}\label{sec10}
The authors declare no competing interests.

\section{Funding}\label{sec11}
This work is funded by the Deutsche Forschungsgemeinschaft (DFG, German Research Foundation) under Germany’s Excellence Strategy\- Cluster of Excellence 2186 "The Integrated Fuel \& Chemical Science Center"\- ID:390919832".

\bibliography{sn-bibliography}

\end{document}